\documentclass[sigconf]{acmart}
\AtBeginDocument{%
  \providecommand\BibTeX{{%
    \normalfont B\kern-0.5em{\scshape i\kern-0.25em b}\kern-0.8em\TeX}}}
    
\copyrightyear{2021}
\acmYear{2021}
\setcopyright{acmlicensed}
\acmConference[MM '21]{Proceedings of the 29th ACM International Conference on Multimedia}{October 20--24, 2021}{Virtual Event, China}
\acmBooktitle{Proceedings of the 29th ACM International Conference on Multimedia (MM '21), October 20--24, 2021, Virtual Event, China}
\acmPrice{15.00}
\acmDOI{10.1145/3474085.3475276}
\acmISBN{978-1-4503-8651-7/21/10}

\usepackage{multicol}
\usepackage{balance}
\usepackage{multirow}
\usepackage{subfigure}
\usepackage{algorithm}
\usepackage{algorithmic}
\usepackage{amsmath}
\usepackage{subfigure}

\begin{document}
\fancyhead{}
\title{Attention-driven Graph Clustering Network}

\settopmatter{authorsperrow=4}

\author{Zhihao Peng}
\affiliation{
  \institution{\small City University of Hong Kong}
  \country{Hong Kong SAR}
  }
\email{zhihapeng3-c@my.cityu.edu.hk}

\author{Hui Liu}
\affiliation{
  \institution{\small City University of Hong Kong}
  \country{Hong Kong SAR}
  }
\email{hliu99-c@my.cityu.edu.hk}

\author{Yuheng Jia}
\authornotemark[1]
\affiliation{
  \institution{\small Southeast University}
  \country{China}
  }
\email{yhjia@seu.edu.cn}

\author{Junhui Hou}
\affiliation{
  \institution{\small City University of Hong Kong}
  \country{Hong Kong SAR}
  }
\email{jh.hou@cityu.edu.hk}

\authornote{\authornotemark[1]Yuheng Jia and Junhui Hou are the corresponding authors. This work was supported by the Hong Kong Research Grants Council under Grant CityU 11219019.}

\begin{abstract}
The combination of the traditional convolutional network (i.e., an auto-encoder) and the graph convolutional network has attracted much attention in clustering, in which the auto-encoder extracts the node attribute feature and the graph convolutional network captures the topological graph feature. However, the existing works 
($i$) lack a flexible combination mechanism to adaptively fuse those two kinds of features for learning the discriminative representation 
and ($ii$) overlook the multi-scale information embedded at different layers for subsequent cluster assignment, 
leading to inferior clustering results. 
To this end, we propose a novel deep clustering method named Attention-driven Graph Clustering Network (AGCN). 
Specifically, AGCN exploits a heterogeneity-wise fusion module to dynamically fuse the node attribute feature and the topological graph feature. Moreover, AGCN develops a scale-wise fusion module to adaptively aggregate the multi-scale features embedded at different layers. Based on a unified optimization framework, AGCN can jointly perform feature learning and cluster assignment in an unsupervised fashion. 
Compared with the existing deep clustering methods, our method is more flexible and effective since it comprehensively considers the numerous and discriminative information embedded in the network and directly produces the clustering results. 
Extensive quantitative and qualitative results on commonly used benchmark datasets validate that our AGCN consistently outperforms state-of-the-art methods.
\end{abstract}




\maketitle

\section{Introduction}

Clustering is a primary yet challenging task in data analysis, aiming to partition similar samples into the same group and dissimilar samples into different groups. Recently, benefiting from the breakthroughs in deep learning, numerous deep clustering approaches have achieved state-of-the-art performance in many applications, including anomaly detection \cite{markovitz2020graph,wang2020cluster,chang2020clustering}, signal propagation \cite{liu2019imbalance,jia2020pairwise,jia2020constrained,huang2021combining,Jia_Liu_Hou_Zhang_2021,jia2021multi}, and transfer clustering \cite{shi2018transfer,peng2019active,han2019learning,peng2020non}.
The crucial prerequisite of deep clustering is to extract intricate patterns from underlying data for effectively learning the data representation. 
For example, Hinton \emph{et al.} \cite{hinton2006reducing} drove the representation learning by a designed auto-encoder network (AE). Xie \emph{et al.} \cite{xie2016unsupervised} proposed the deep embedded clustering method (DEC) to learn the feature representation by clustering a set of data points in a jointly optimized feature space. Guo \emph{et al.} \cite{guo2017improved} introduced a reconstruction loss to improve DEC for learning a better representation. Although these works have achieved remarkable improvements, they simply focus on the node attribute feature and ignore the topological graph information embedded in the data.

As the topological graph information can make a valuable guide on embedding learning, various works \cite{kipf2016variational,velivckovic2018graph,park2019symmetric,8822591,bo2020structural} have been proposed to introduce the graph convolutional networks (GCNs) to use the topological graph information for learning the graph structure feature. Specifically, Kipf \emph{et al.} \cite{kipf2016variational} proposed the graph auto-encoder (GAE) and the variational graph auto-encoder (VGAE) to learn the graph structure feature based on the AE and the variational AE based frameworks, respectively. Furthermore, based on the GAE framework, Pan \emph{et al.} \cite{8822591} developed the adversarially regularized graph auto-encoder network (ARGA) by introducing an adversarial regularizer. Wang \emph{et al.} \cite{wang2019attributed} combined GAE with the graph attention network model \cite{velivckovic2018graph} to encode the topological structure and node contents. Bo \emph{et al.} \cite{bo2020structural} designed the structural deep clustering network (SDCN) to integrate the topological graph information into deep clustering based on the DEC framework. However, these existing works naively equate the importance of the topological graph feature and the node attribute feature in any case, inevitably limiting the representation learning. Moreover, they only consider the latent features extracted from the deepest layer, neglecting the off-the-shelf yet discriminative multi-scale information embedded in different layers.

In this paper, we propose a novel deep clustering method named attention-driven graph clustering network (AGCN) to address the above-mentioned issues. Specifically, AGCN includes two fusion modules, namely AGCN heterogeneity-wise\footnote{Here, `heterogeneity' indicates the discrimination of feature structure, e.g., the GCN-based feature structure and the AE-based feature structure.} fusion module (AGCN-H) and AGCN scale-wise fusion module (AGCN-S), in which both modules exploit the attention-based mechanism to dynamically measure the importance of the corresponding features for the subsequent feature fusion. AGCN-H adaptively merges the GCN feature and the AE feature from the same layer, while AGCN-S dynamically concatenates the multi-scale features from different layers. For conducting the training process in an unsupervised fashion, we design a unified learning framework capable of directly producing the clustering results. Extensive quantitative and qualitative comparisons are conducted on six commonly used benchmark datasets to validate the superiority of AGCN over state-of-the-art methods. Furthermore, the ablation studies are performed to validate the efficiency and effectiveness of our approach.

\textbf{Notation:}
Throughout this paper, scalars are denoted by italic lower case letters, vectors by bold lower case letters, matrices by upper case ones, and operators by calligraphy ones, respectively. Let $\mathbf{V}$ be the set of nodes, $\mathbf{E}$ be the set of edges between nodes, $\mathbf{X} \in\mathbb{R}^{n\times d}$ be the node attribute matrix, then $G=(\mathbf{V},\mathbf{E},\mathbf{X})$ denotes the undirected graph. The adjacency matrix $\mathbf{A} \in\mathbb{R}^{n\times n}$ indicates the topological structure of graph $\emph{G}$ and the corresponding degree matrix is $\mathbf{D} \in\mathbb{R}^{n\times n}$. $\left\|\cdot\right\|_F$ denotes the Frobenius norm. The main notations used throughout the paper are summarized in Table \ref{tab: notation}.

\section{Related work}

Recently, many deep clustering methods \cite{xie2016unsupervised,guo2017improved,han2019learning,peng2021maximum,li2020deep,affeldt2020spectral} have been proposed and achieved impressive performance, benefiting from the strong representation power of the deep neural networks. Auto-encoder (AE) \cite{hinton2006reducing} is one of the most commonly used unsupervised deep neural networks, which plays a crucial role in deep clustering. For example, the deep embedded clustering (DEC) \cite{xie2016unsupervised} used the AE-based framework to learn the deep representations by Kullback-Leibler (KL) divergence minimization. The improved DEC method (IDEC) \cite{guo2017improved} promoted the clustering performance of DEC. \cite{han2019learning} achieved the deep transfer clustering by simultaneously learning the data representation and clustering the unlabelled data of novel visual categories. \cite{li2020deep} incorporated adversarial fairness to complete the group invariant cluster assignment and the structural preservation. However, these methods only focus on learning the data representation from the samples themselves and overlook the potential valuable graph structure information between data samples. 

To exploit the structural information underlying the data, some graph convolutional networks (GCNs) based clustering methods were proposed \cite{kipf2016variational,velivckovic2018graph,wu2020comprehensive,park2019symmetric,wang2019attributed,wang2020gcn,8822591,TuDeep,kim2021find,zhu2021deep}. For instance, \cite{kipf2016variational} proposed using the graph auto-encoder (GAE) and the variational graph auto-encoder (VGAE) to learn the graph-structured data. \cite{wang2019attributed} provided the deep attentional embedded graph clustering network (DAEGC) to encode the topological structure and node contents in a graph by introducing the attentional neighbor-wise fusion strategy on the GAE framework. The adversarially regularized graph auto-encoder (ARGA) \cite{8822591} further improved the clustering performance by introducing an adversarial learning scheme to learn the graph embedding. \cite{bo2020structural} designed the structural deep clustering network (SDCN) to integrate the structural information into deep clustering by embedding GCN into the DEC framework. 

\begin{table}[ht]
\setlength{\abovecaptionskip}{2pt}    
\setlength{\abovecaptionskip}{8pt}
    \caption{Main notations and descriptions.}
    \label{tab: notation}
    \centering
        \begin{tabular}{l|l}
        \hline\hline
        Notations & Descriptions                \\ \hline
        $\mathbf{X,\hat{\mathbf{X}}}$  & The input data and its reconstructed matrix  \\
        $\mathbf{H}$            & The extracted feature from AE module  \\
        $\mathbf{A}, \mathbf{D}$            & The adjacency matrix and the degree matrix\\
        $\mathbf{Z}_i$, $\mathbf{H}_i$          & The GCN and encoder output from the $i_{th}$ layer \\
        $\mathbf{M}_i$          & The AGCN-H weight matrix for $\mathbf{Z}_i$ and $\mathbf{H}_i$ \\
        $\mathbf{Z}_i^{'}$      & The AGCN-H combined feature for $\mathbf{Z}_i$ and $\mathbf{H}_i$\\
        $\mathbf{U}$, $\mathbf{u}_i$            & The AGCN-S weight matrix and its elements \\
        $\mathbf{Z}^{'}$        & The AGCN-S combined feature \\
        $\mathbf{Z}$            & The soft assignment   \\
        \hline
        $n$, $l$, $\emph{k}$                      & The number of samples, network layers, and clusters  \\
        $d$, $d_i$                     & The dimension of $\mathbf{X}$ and the $i_{th}$ latent feature  \\
        \hline
        $\cdot \| \cdot $       & The concatenation operation \\
    $\left\|\cdot\right\|_F$    & The Frobenius norm \\
        \hline\hline   
        \end{tabular}
\end{table}

Although the above approaches can improve the clustering performance, they still have the following drawbacks, i.e., ($i$) naively equating the importance of the topological graph feature and the node attribute feature; ($ii$) neglecting the multi-scale information embedded in different layers. Accordingly, embedding learning cannot effectively and comprehensively exploit the graph structure of data. Moreover, the interaction between the graph structure feature and the node attribute feature is not adequate to a certain extent. As a result, the fruitful and valuable information is dropping, limiting the performance of the clustering model.

\begin{figure*}%[h]
	\centering
	\includegraphics [width=1.66\columnwidth]{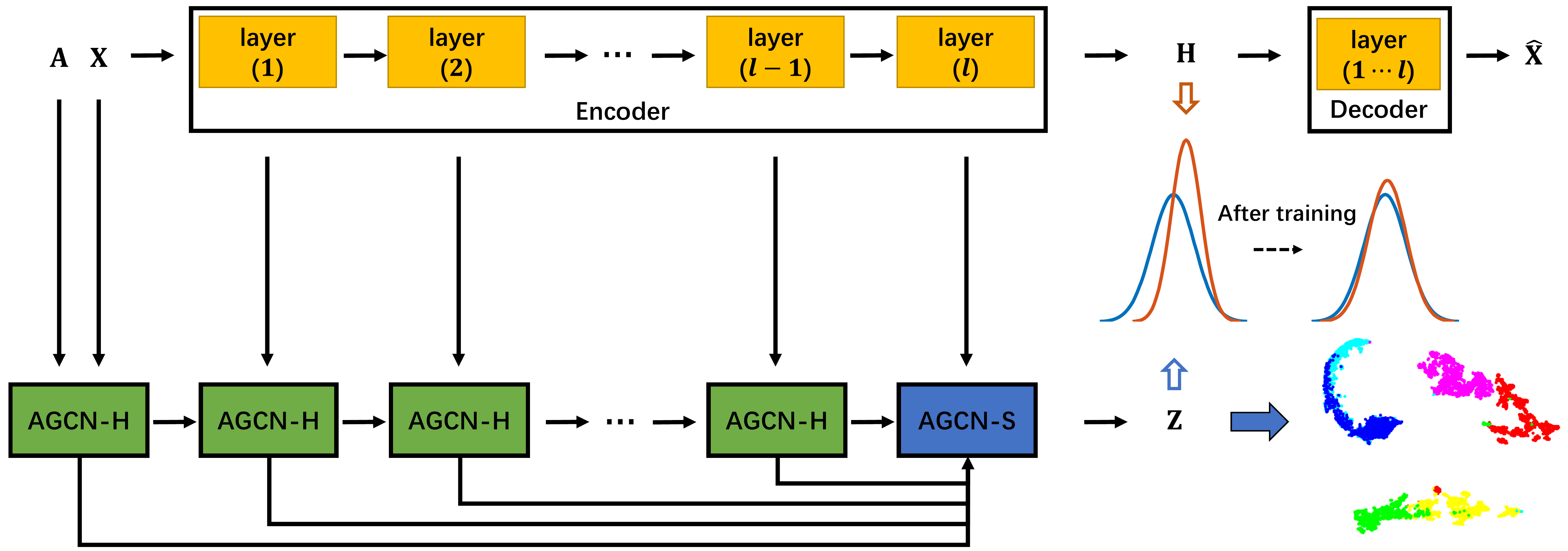}
    \setlength{\abovecaptionskip}{2pt}    
    \setlength{\abovecaptionskip}{8pt}
	\caption{
	The architecture of the proposed attention-driven graph clustering network (AGCN). $\mathbf{X}$ denotes the input data, $\mathbf{A}$ denotes the adjacency matrix, $\mathbf{\hat{X}}$ denotes the reconstructed data, $\mathbf{\emph{l}}$ denotes the number of layers. The upper part is an encoder-decoder (i.e., auto-encoder) module that the latent representation $\mathbf{H}$ is extracted by minimizing the reconstruction loss between $\mathbf{X}$ and $\mathbf{\hat{X}}$. The lower part consists of the proposed AGCN heterogeneity-wise fusion module (AGCN-H) and scale-wise fusion module (AGCN-S), in which AGCN-H and AGCN-S are designed to achieve the heterogeneous features fusion and the multi-scale features fusion, respectively. The network is self-trained by minimizing the KL divergence between the $\mathbf{H}$ distribution (as indicated in orange) and the $\mathbf{Z}$ distribution (as indicated in blue).}
	\vspace{-0.3cm}
	\label{fig: Our_framework}
\end{figure*}

\section{Proposed method}

In this section, we first describe the details of the proposed attention-driven graph clustering network (AGCN) shown in Figure \ref{fig: Our_framework}, including the heterogeneity-wise fusion module (AGCN-H) and the scale-wise fusion module (AGCN-S). Then, we introduce the network training process and the computational complexity analysis.

\subsection{AGCN-H}

\begin{figure*}%[h]
	\centering
    \includegraphics [width=0.98\textwidth]{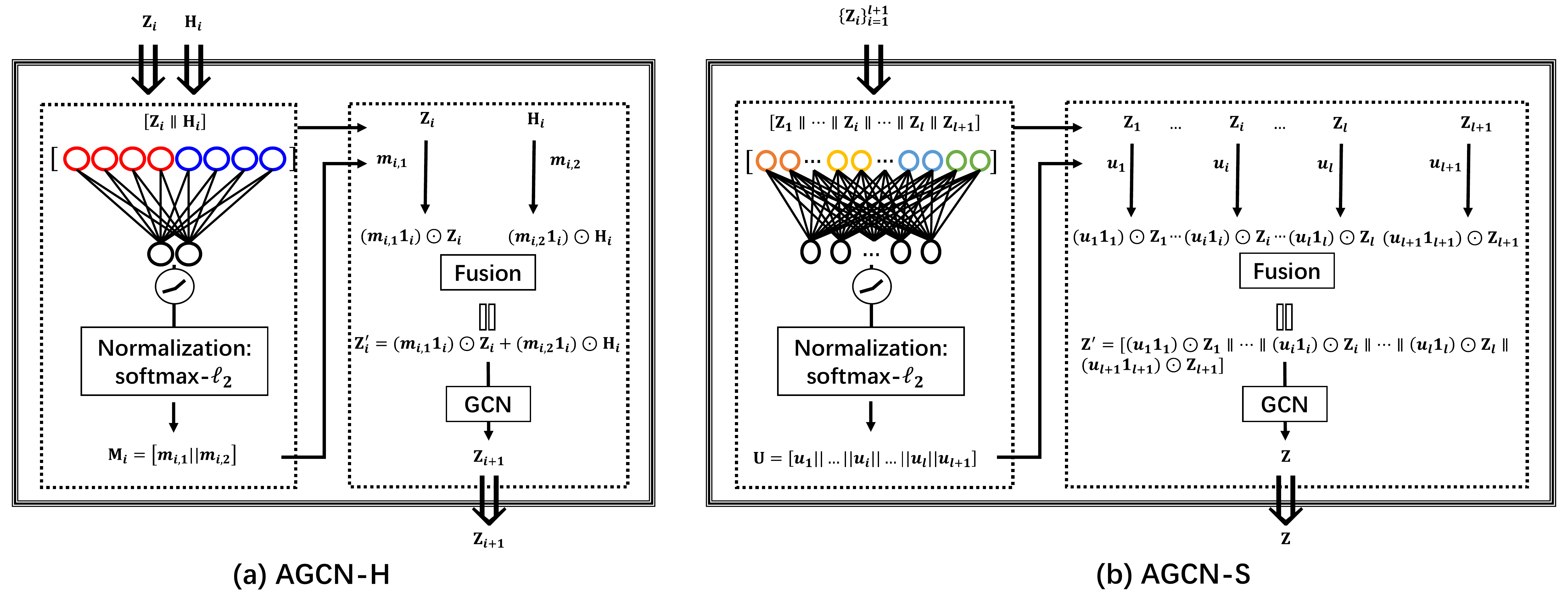}
    \setlength{\abovecaptionskip}{2pt}%    
    \setlength{\abovecaptionskip}{8pt}%
    \caption{The illustrations of the proposed (a) AGCN-H module and (b) AGCN-S module. In (a), the GCN feature $\mathbf{Z}_{i}$ and the AE feature $\mathbf{H}_{i}$ are fused to obtain $\mathbf{Z}_{i+1}$ via a weighted sum form. In (b), the multi-scale weighted features are combined in a feature concatenation manner. Specifically, we first learn the weights through the proposed attention-based mechanism (the left dashed box in the triple-solid line box) and then integrate the corresponding features through the weighted fusion (the right dashed box in the triple-solid line box). Here, the `softmax-$\ell_{2}$' normalization means using the softmax function and $\ell_{2}$ normalization. The input and output actions of the modules are represented by $\Downarrow$.}
    \vspace{-0.3cm}
	\label{fig: AGCN-H-S}
\end{figure*}

As the graph convolutional network (GCN) can efficiently capture the topological graph information and the auto-encoder (AE) can reasonably extract the node attribute feature, we propose the AGCN-H module to dynamically combine the GCN feature and the AE feature to learn a more discriminative representation. Specifically, we exploit the attention-based mechanism with the heterogeneity-wise strategy by conducting the attention coefficients learning and the subsequent weighted feature fusion. The corresponding illustration of AGCN-H is shown in Figure \ref{fig: AGCN-H-S} (a), and the implementation details are as follows.

First, the encoder-decoder module is used to extract the latent representation by minimizing the reconstruction loss between the raw data and the reconstructed data, i.e.,

\begin{equation}
\begin{aligned}
&\mathcal{L}_{R} = \left\| \mathbf{X} - \hat{\mathbf{X}} \right\|^2_F\\
& \rm{s.t.} \quad \{ \mathbf{H}_{\emph{i}} = \phi ( \mathbf{W}_{\emph{i}}^{e}\mathbf{H}_{\emph{i}-1}+ \mathbf{b}_{\emph{i}}^{e}), \\ 
& \textcolor{black}{\hat{\mathbf{H}}_{\emph{i}} = \phi ( \mathbf{W}_{\emph{i}}^{d}\hat{\mathbf{H}}_{\emph{i}-1}+ \mathbf{b}_{\emph{i}}^{d}),} \emph{i}={1,\cdots,\emph{l} } \},
\label{eq: AE}
\end{aligned}
\end{equation}
\textcolor{black}{where $\mathbf{X}\in\mathbb{R}^{n\times d}$ denotes the raw data, $\hat{\mathbf{X}}\in\mathbb{R}^{n\times d}$ denotes the reconstructed data, $\mathbf{H}_{\emph{i}}\in\mathbb{R}^{n\times {d_i}}$ and $\hat{\mathbf{H}}_{\emph{i}}\in\mathbb{R}^{n\times {\hat{d}_i}}$ denote the encoder and decoder outputs from the $i_{th}$ layer, respectively. $\phi ( \cdot )$ denotes the activation function such as Tanh, ReLU \cite{glorot2011deep}, etc. $\mathbf{W}_{\emph{i}}^{e}$ and $\mathbf{b}_{i}^{e}$ denote the network weight and bias of the $i_{th}$ encoder layer, respectively. $\mathbf{W}_{\emph{i}}^{d}$ and $\mathbf{b}_{i}^{d}$ denote the network weight and bias of the $i_{th}$ decoder layer, respectively. Particularly, $\mathbf{H}_{0}$ indicates the raw data $\mathbf{X}$ and $\hat{\mathbf{H}}_{l}$ indicates the reconstructed data $\hat{\mathbf{X}}$.} In addition, let the GCN feature learned from the $i_{th}$ layer be $\mathbf{Z}_i\in\mathbb{R}^{n\times d_i}$, where $\mathbf{Z}_{0}$ indicates the raw data $\mathbf{X}$.

To learn the corresponding attention coefficients, $\mathbf{Z}_i$ and $\mathbf{H}_i$ are first concatenated as $[\mathbf{Z}_\emph{i}\|\mathbf{H}_\emph{i}]\in\mathbb{R}^{n\times {2d_i}}$. Then, a full-connected layer, parametrized by a weight matrix $\mathbf{W}_{\emph{i}}^{a}\in\mathbb{R}^{{2d_i}\times 2}$, is introduced to capture the relationship for the concatenated features. Afterwards, the LeakyReLU \cite{maas2013rectifier} activation function (negative input slope is set as $0.2$) is applied on the multiplication between $\left[\mathbf{Z}_\emph{i}\|\mathbf{H}_\emph{i}\right]$ and $\mathbf{W}_{\emph{i}}^{a}$. We then normalize the output of the LeakyReLU unit via the softmax function and the $\ell_2$ normalization (indicated as the `softmax-$\ell_{2}$' normalization). The corresponding expression is formulated as
\begin{equation}
\begin{aligned}
\mathbf{M}_\emph{i}=\ell_{2}\left( softmax \left( \left(LeakyReLU\left(\left[\mathbf{Z}_\emph{i}\|\mathbf{H}_\emph{i}\right]\mathbf{W}_{\emph{i}}^{a}\right)\right) \right) \right),
\label{eq: AGCN-H-A}
\end{aligned}
\end{equation}
where $\mathbf{M}_\emph{i}=[\mathbf{m}_{i,1}\| \mathbf{m}_{i,2}]\in\mathbb{R}^{n\times 2}$ is the attention coefficient matrix with entries being greater than $0$, and $\mathbf{m}_{i,1},\mathbf{m}_{i,2}$ are the weight vectors for measuring the importance of $\mathbf{Z}_i$ and $\mathbf{H}_i$, respectively. Accordingly, we adaptively fuse the GCN feature $\mathbf{Z}_i$ and the AE feature $\mathbf{H}_i$ on the $i_{th}$ layer as,

\begin{equation}
\begin{aligned}
& \mathbf{Z}_\emph{i}^{'}= \left( \mathbf{m}_{i,1}\mathbf{1}_i \right) \odot \mathbf{Z}_\emph{i} + \left( \mathbf{m}_{i,2}\mathbf{1}_i \right) \odot \mathbf{H}_\emph{i},
\label{eq: AGCN-H}
\end{aligned}
\end{equation}
where $\mathbf{1}_{i}\in\mathbb{R}^{1\times d_{i}}$ denotes the vector of all ones, `$\odot$' denotes the Hadamard product of matrices. Then, the obtained matrix $\mathbf{Z}_{i}^{'}\in\mathbb{R}^{n\times d_{i}}$ is used as the input of the $(\emph{i}+1)_{th}$ GCN layer to learn the representation $\mathbf{Z}_{\emph{i}+1}$, which can be formulated as
\begin{equation}
\begin{aligned}
& \mathbf{Z}_{\emph{i}+1}= LeakyReLU(\mathbf{D}^{-\frac{1}{2}}(\mathbf{A}+\mathbf{I})\mathbf{D}^{-\frac{1}{2}}\mathbf{Z}_\emph{i}^{'}\mathbf{W}_\emph{i}),
\label{eq: AGCN-H-output}
\end{aligned}
\end{equation}
where the original adjacency matrix $\mathbf{A}$ is normalized via $\mathbf{D}^{-\frac{1}{2}}(\mathbf{A}+\mathbf{I})\mathbf{D}^{-\frac{1}{2}}$ with $\mathbf{I}\in\mathbb{R}^{n\times n}$ being the identity matrix, $\mathbf{D}$ being the corresponding degree matrix, $\mathbf{W}_\emph{i}$ denotes the network weight. In summary, we are capable of achieving the dynamic feature fusion between the GCN and AE features through the AGCN-H module.

\subsection{AGCN-S}
Considering that the current deep clustering algorithms usually neglect the multi-scale information embedded in different layers, we thus design the AGCN-S module to exploit the multi-scale information. As the dimensions of the features at different layers are different, we preliminarily aggregate the multi-scale features with a concatenation manner, which is formulated as
\begin{equation}
\begin{aligned}
& \mathbf{Z}^{'} = \left[\mathbf{Z}_1\|\cdots\|\mathbf{Z}_{\emph{i}}\|\cdots\|\mathbf{Z}_{\emph{l}}\|\mathbf{Z}_{\emph{l+1}}\right],
\label{eq: AGCN-S-S}
\end{aligned}
\end{equation}
where $\mathbf{Z}_i\in\mathbb{R}^{n\times d_i}$ with $d_{i}$ being the dimension of the $i_{th}$ layer, $l$ denotes the number of encoder layers. Particularly, $\mathbf{Z}_{\emph{l+1}}=\mathbf{H}_l\in\mathbb{R}^{n\times d_l}$. 

Motivated by the fact that the features at different layers depict the input data with different levels of semantic description, and accordingly may play different roles in the final clustering task, naively equating the importance of different scale features in feature fusion is not desirable. To this end, we develop the AGCN-S module to dynamically combine various scale features via the attention-based mechanism. The corresponding illustration is shown in Figure \ref{fig: AGCN-H-S} (b), and the implementation details are as follows.

First, we use a full-connected layer, parametrized by a weight matrix $\mathbf{W}^{s}\in\mathbb{R}^{(d_1+\cdots+d_l+d_l) \times (\emph{l}+1)}$ to capture the relationship among the features at different layers, and apply the LeakyReLU activation function on the multiplication between $\left[\mathbf{Z}_1\|\cdots\|\mathbf{Z}_{\emph{i}}\|\cdots\|\mathbf{Z}_{\emph{l}}\|\mathbf{Z}_{\emph{l+1}}\right]$ and $\mathbf{W}^{s}$. After that, by using the `softmax-$\ell_2$' normalization on each row's elements, we normalize them to scale the output weight value for making the attention coefficients easily comparable. Technically, the attention coefficient matrix can be expressed as
\begin{equation}
\begin{aligned}
\mathbf{U}=\ell_{2}( softmax( LeakyReLU( [\mathbf{Z}_1\|\cdots\|\mathbf{Z}_{\emph{i}}\|\cdots\|\mathbf{Z}_{\emph{l}}\|\mathbf{Z}_{\emph{l+1}}]\mathbf{W}^{s}) ) ),
\label{eq: AGCN-S-A}
\end{aligned}
\end{equation}
where $\mathbf{U}= \left[\mathbf{u}_1\|\cdots\|\mathbf{u}_{\emph{i}}\|\cdots\|\mathbf{u}_{\emph{l}}\|\mathbf{u}_{\emph{l+1}}\right]\in\mathbb{R}^{n\times {(l+1)}}$ with entries being greater than $0$, $\mathbf{u}_{\emph{i}}$ being the parallel attention coefficient for $\mathbf{Z}_i$. 

To sufficiently explore the information embedded on multi-scale features, we then impose the attention-based scale-wise strategy to Eq. (\ref{eq: AGCN-S-S}), i.e., weighting the multi-scale features with the learned attention coefficients. In this way, the feature fusion can be formulated as
\begin{equation}
\begin{aligned}
\mathbf{Z}^{'} = 
&[\left(\mathbf{u}_{1}\mathbf{1}_{1}\right)\odot\mathbf{Z}_1\|\cdots\|\left(\mathbf{u}_{i}\mathbf{1}_{i}\right)\odot\mathbf{Z}_{\emph{i}}\|\cdots\|\left(\mathbf{u}_{l}\mathbf{1}_{l}\right)\odot\mathbf{Z}_{\emph{l}}\|\\
&\left(\mathbf{u}_{l+1}\mathbf{1}_{l+1}\right)\odot\mathbf{Z}_{\emph{l}+1}].
\label{eq: AGCN-S}
\end{aligned}
\end{equation}
The fused feature $\mathbf{Z}^{'}$ is used as the input of the final prediction layer to learn the representation $\mathbf{Z}\in \mathbb{R}^{n\times \emph{k}}$ with $\emph{k}$ being the cluster number. A Laplacian smoothing operator \cite{li2018deeper} and a softmax function are used to obtain the reasonable probability distribution for subsequent prediction, which is as follows:
\begin{equation}
\begin{aligned}
& \mathbf{Z}= softmax(\mathbf{D}^{-\frac{1}{2}}(\mathbf{A}+\mathbf{I})\mathbf{D}^{-\frac{1}{2}}\mathbf{Z}^{'}\mathbf{W}) \\
& \rm{s.t.} \quad \sum_{\emph{j}=1}^{\emph{k}}\emph{z}_{\emph{i},\emph{j}}=1, \emph{z}_{\emph{i},\emph{j}}>0,
\label{eq: AGCN-output}
\end{aligned}
\end{equation}
where $\textbf{W}$ denotes the learnable parameters. When the network is well-trained, we can directly infer the predicted cluster label through $\mathbf{Z}$, i.e.,
\begin{equation}
\begin{aligned}
& y_{i}=\mathop{\arg\max}_{\emph{j}} \mathbf{z}_{i,\emph{j}} \\ 
& \rm{s.t.} \quad \emph{j}=1,\cdots,\emph{k},
\label{eq: clustering_result}
\end{aligned}
\end{equation}
where $y_{i}$ is the predicted label of data $\textbf{x}_{i}$.

\subsection{Training process}
As clustering is an unsupervised task without reliable guidance, it is crucial to exploit the relationship between the AE feature and the combined feature to drive the network training. To this end, we unify the AE feature and the combined feature in a uniform framework, and a practical end-to-end solution is designed for network training. The training process includes two steps:

$\textbf{\emph{Step 1.}}$ \textcolor{black}{To adopt the learned features of our method to the clustering task, we used the Student’s t-distribution \cite{helmert1876genauigkeit,student1908probable} as a kernel to measure the similarity between embedded point and centroid, in which the measured similarity can be interpreted as the soft assignment. After that, our model can iteratively refine clusters with an auxiliary target distribution derived from the current soft assignment, which is a commonly used strategy to achieve clustering in many recent deep clustering methods \cite{xie2016unsupervised,jabi2019deep,li2020deep}. The formulation is as follows, }
\begin{equation}
\begin{aligned}
&q_{i,\emph{j}} = \frac{(1+\|\mathbf{h}_{i}-\mathbf{\mu}_{\emph{j}}\|^2/\alpha)^{-\frac{\alpha+1}{2}}}{ \sum_{\emph{j}^{'}} (1+\|\mathbf{h}_{i}-\mathbf{\mu}_{\emph{j}^{'}}\|^2/\alpha)^{-\frac{\alpha+1}{2}} },\\
\label{eq: KL-q}
\end{aligned}
\end{equation}
where $\mathbf{H}=\mathbf{H}_{l}=[\mathbf{h}_1,\cdots,\mathbf{h}_{n}]^\mathsf{T}$, $q_{i,\emph{j}}$ denotes the similarity between $\mathbf{h}_{i}$ and its corresponding cluster center vector $\mathbf{\mu}_j$, $\alpha$ is set to $1$. As directly minimizing the KL divergence between distributions of $\mathbf{Z}$ and $\mathbf{H}$ may bring trivial solutions \cite{bo2020structural}, we introduce an auxiliary target distribution $\mathbf{P}$ to avoid the collapse issue, i.e.,
\begin{equation}
\begin{aligned}
&p_{i,\emph{j}}=\frac{ q_{i,\emph{j}}^{2}/\sum_{i}  q_{i,\emph{j}} }{\sum_{\emph{j}}^{'} q_{i,\emph{j}^{'}}^{2}/\sum_{i}  q_{i,\emph{j}^{'}}},
\label{eq: KL-p}
\end{aligned}
\end{equation}
where $0\leq p_{i,\emph{j}}\leq 1$ is the element of $\mathbf{P}$.

$\textbf{\emph{Step 2.}}$ We minimize the KL divergence between the combined feature $\mathbf{Z}$ distribution and the AE feature $\mathbf{H}$ distribution with the help of the auxiliary target distribution $\mathbf{P}$, which can be formulated as
\begin{equation}
\begin{aligned}
\mathcal{L}_{KL} 
&= \lambda_1*KL(\mathbf{P}, \mathbf{Z}) + \lambda_2*KL(\mathbf{P}, \mathbf{H}) \\
&=\lambda_1\sum_i\sum_j{p_{i,\emph{j}} log{\frac{p_{i,\emph{j}}}{z_{i,\emph{j}}}}} + \lambda_2\sum_i\sum_j{p_{i,\emph{j}} log{\frac{p_{i,\emph{j}}}{q_{i,\emph{j}}}}},\\
\label{eq: KL}
\end{aligned}
\end{equation}
where $\lambda_1>0$ and $\lambda_2>0$ are the trade-off parameters. By minimizing Eq. (\ref{eq: KL}), the distributions of $\mathbf{Z}$ and $\mathbf{H}$ can be well aligned. Combining the Eq. (\ref{eq: AE}) and Eq. (\ref{eq: KL}), the overall loss function of our method can be written as
\begin{equation}
\begin{aligned}
&\mathcal{L} = \mathcal{L}_{R}+\mathcal{L}_{KL},\\
\label{eq: ASCN_loss}
\end{aligned}
\end{equation}
where $\mathcal{L}_{R}$ is the reconstruction loss of AE, $\mathcal{L}_{KL}$ is the alignment loss with the combined feature $\mathbf{Z}$ and the AE feature $\mathbf{H}$. The training process of our method AGCN is shown in Algorithm \ref{alg1}. 

\begin{algorithm}
	\caption{Training process of AGCN}
	\label{alg1}
	\begin{algorithmic}
		\REQUIRE Input data $\mathbf{X}$; Adjacency matrix $\mathbf{A}$; Cluster number $\emph{k}$; \textcolor{black}{Network layers number $l=4$}; Trade-off parameters $\lambda_1,\lambda_2$; Maximum iterations $i_\emph{MaxIter}$;\\
		\ENSURE  Clustering result $\mathbf{y}$;\\
	\end{algorithmic}
	\begin{algorithmic}[1]
    	\STATE Initialization: $i_\emph{Iter} = 1$; $\mathbf{Z}_0 = \mathbf{X}$; $\mathbf{H}_0 = \mathbf{X}$;
    	\STATE Initialize the parameters of auto-encoder;
		\WHILE{$i_\emph{Iter} < i_\emph{MaxIter}$}
		\STATE Obtain the AE feature $\mathbf{H}$ by Eq. (\ref{eq: AE});
		\STATE Obtain the fused features of AGCN-H module via Eq. (\ref{eq: AGCN-H-output});
		\STATE Obtain the fused features of AGCN-S module via Eq. (\ref{eq: AGCN-S});
		\STATE Obtain the combined feature $\mathbf{Z}$ via Eq. (\ref{eq: AGCN-output});
		\STATE Obtain the cluster center embedding $\boldsymbol{\mu}$ with K-means based on the feature $\mathbf{H}$;
		\STATE Use the feature $\mathbf{H}$ and cluster center embedding $\boldsymbol{\mu}$ to calculate the AE feature distribution $\mathbf{H}$ via Eq. (\ref{eq: KL-q});
		\STATE Calculate the auxiliary target distribution $\mathbf{P}$ via Eq. (\ref{eq: KL-p});
		\STATE Minimize the KL divergence between distribution $\mathbf{Z}$ and distribution $\mathbf{H}$ via Eq. (\ref{eq: KL});
		\STATE Calculate $\mathcal{L}_{R}$ and $\mathcal{L}_{KL}$ respectively;
		\STATE Calculate the overall loss function via Eq. (\ref{eq: ASCN_loss});
		\STATE Conduct the back propagation and update parameters in the proposed AGCN network;
		\STATE $i_\emph{Iter} = i_\emph{Iter} + 1$;
		\ENDWHILE \\
		\STATE Calculate the clustering results $\mathbf{y}$ with the combined feature $\mathbf{Z}$ by Eq. (\ref{eq: clustering_result});
	\end{algorithmic}
\end{algorithm}	

\subsection{Computational Complexity Analysis}
Given $n$ being the number of samples, $d$ being the dimension of input data, $d_i$ being the dimension of the $i_{th}$ layer, $l$ being the number of layers, and $\emph{k}$ being the number of clusters. For the auto-encoder, the time complexity is $\mathcal{O}_{1}=\mathcal{O}(n\sum_{i=2}^{l}d_{i-1}d_{i})$. For the GCN module, as the operation can be computed efficiently using sparse matrix computation, the time complexity is $\mathcal{O}_{2}=\mathcal{O}(|\mathbf{E}|\sum_{i=2}^{l}d_{i-1}d_{i})$ corresponding to \cite{8822591}. For Eq. (\ref{eq: KL-q}), the time complexity is $\mathcal{O}_{3}=\mathcal{O}(nk+n\log n)$ corresponding to \cite{xie2016unsupervised}. For our proposed modules, the time complexity is $\mathcal{O}_{4}=\mathcal{O}(\sum_{i=1}^{l-1}(d_{i}))+\mathcal{O}((\sum_{i=1}^{l+1} d_{i})(l+1))$ with $d_{l+1}=d_{l}$. \textcolor{black}{Thus, the total computational complexity of Algorithm \ref{alg1} in one iteration is about $\mathcal{O}(n\sum_{i=2}^{l}d_{i-1}d_{i} + |\mathbf{E}|\sum_{i=2}^{l}d_{i-1}d_{i} + nk+n\log n + \sum_{i=1}^{l-1}(d_{i}) + (\sum_{i=1}^{l+1} d_{i})(l+1))$.}

\section{Experiments}
\subsection{Datasets}

We conduct the experiments on six commonly used benchmark datasets, including one image dataset (USPS \cite{hull1994database}), one record dataset (HHAR \cite{stisen2015smart}), one text dataset (Reuters \cite{lewis2004rcv1}), and three graph datasets (ACM$\footnote{http://dl.acm.org}$, CiteSeer$\footnote{http://CiteSeerx.ist.psu.edu/}$, and  DBLP$\footnote{https://dblp.uni-trier.de}$). 

\begin{table}[]
\setlength{\abovecaptionskip}{2pt}%    
\setlength{\abovecaptionskip}{8pt}%
\caption{Description of the adopted datasets.}
\label{tab: datasets}
\begin{tabular}{l|c|c|c|c}
\hline\hline
Dataset  & Type   & Samples & Classes & Dimension \\ \hline
USPS     & Image  & 9298    & 10      & 256       \\
HHAR     & Record & 10299   & 6       & 561       \\
Reuters  & Text   & 10000   & 4       & 2000      \\
ACM      & Graph  & 3025    & 3       & 1870      \\
CiteSeer & Graph  & 3327    & 6       & 3703      \\ 
DBLP     & Graph  & 4057    & 4       & 334       \\ \hline\hline
\end{tabular}
\end{table}

The brief of the used datasets is summarized in Table \ref{tab: datasets}. For the non-graph data, the adjacency matrix $\mathbf{A}$ is generated by the undirected $\emph{k'}$-nearest neighbor (KNN \cite{altman1992introduction}) graph following \cite{bo2020structural}.

\begin{table*}[]
\setlength{\abovecaptionskip}{2pt}%    
\setlength{\abovecaptionskip}{8pt}%
\caption{Clustering performance on six datasets (mean$\pm$std). The best and second-best results are highlighted with \textbf{bold} and \underline{underline}, respectively.}
\label{tab: final_result}
\setlength{\tabcolsep}{1.48mm}{
\begin{tabular}{l|c|cccccccc|c}
\hline\hline
Dataset                   & Metric & AE             & DEC            & IDEC           & GAE            & VGAE           & DAEGC          & ARGA           & SDCN                    & Our                                         \\ \hline
\multirow{4}{*}{USPS}     & ACC    & 71.04$\pm$0.03 & 73.31$\pm$0.17 & 76.22$\pm$0.12 & 63.10$\pm$0.33 & 56.19$\pm$0.72 & 73.55$\pm$0.40 & 66.80$\pm$0.70 & {\underline{ 78.08$\pm$0.19}}    &                            \textbf{80.98$\pm$0.28} \\
                          & NMI    & 67.53$\pm$0.03 & 70.58$\pm$0.25 & 75.56$\pm$0.06 & 60.69$\pm$0.58 & 51.08$\pm$0.37 & 71.12$\pm$0.24 & 61.60$\pm$0.30 & {\underline{ 79.51$\pm$0.27}}    & \textbf{79.64$\pm$0.32} \\
                          & ARI    & 58.83$\pm$0.05 & 63.70$\pm$0.27 & 67.86$\pm$0.12 & 50.30$\pm$0.55 & 40.96$\pm$0.59 & 63.33$\pm$0.34 & 51.10$\pm$0.60 & {\underline{ 71.84$\pm$0.24}}    & \textbf{73.61$\pm$0.43} \\
                          & F1     & 69.74$\pm$0.03 & 71.82$\pm$0.21 & 74.63$\pm$0.10 & 61.84$\pm$0.43 & 53.63$\pm$1.05 & 72.45$\pm$0.49 & 66.10$\pm$1.20 & {\underline{ 76.98$\pm$0.18}}    & \textbf{77.61$\pm$0.38} \\ \hline
\multirow{4}{*}{HHAR}     & ACC    & 68.69$\pm$0.31 & 69.39$\pm$0.25 & 71.05$\pm$0.36 & 62.33$\pm$1.01 & 71.30$\pm$0.36 & 76.51$\pm$2.19 & 63.30$\pm$0.80 & {\underline{ 84.26$\pm$0.17}}    & \textbf{88.11$\pm$0.43}                     \\
                          & NMI    & 71.42$\pm$0.97 & 72.91$\pm$0.39 & 74.19$\pm$0.39 & 55.06$\pm$1.39 & 62.95$\pm$0.36 & 69.10$\pm$2.28 & 57.10$\pm$1.40 & {\underline{ 79.90$\pm$0.09}}    & \textbf{82.44$\pm$0.62}                     \\
                          & ARI    & 60.36$\pm$0.88 & 61.25$\pm$0.51 & 62.83$\pm$0.45 & 42.63$\pm$1.63 & 51.47$\pm$0.73 & 60.38$\pm$2.15 & 44.70$\pm$1.00 & {\underline{ 72.84$\pm$0.09}}    & \textbf{77.07$\pm$0.66}                     \\
                          & F1     & 66.36$\pm$0.34 & 67.29$\pm$0.29 & 68.63$\pm$0.33 & 62.64$\pm$0.97 & 71.55$\pm$0.29 & 76.89$\pm$2.18 & 61.10$\pm$0.90 & {\underline{ 82.58$\pm$0.08}}    & \textbf{88.00$\pm$0.53}                     \\ \hline
\multirow{4}{*}{Reuters}  & ACC    & 74.90$\pm$0.21 & 73.58$\pm$0.13 & 75.43$\pm$0.14 & 54.40$\pm$0.27 & 60.85$\pm$0.23 & 65.50$\pm$0.13 & 56.20$\pm$0.20 & {\underline{ 77.15$\pm$0.21}}    & \textbf{79.30$\pm$1.07}                     \\
                          & NMI    & 49.69$\pm$0.29 & 47.50$\pm$0.34 & 50.28$\pm$0.17 & 25.92$\pm$0.41 & 25.51$\pm$0.22 & 30.55$\pm$0.29 & 28.70$\pm$0.30 & {\underline{ 50.82$\pm$0.21}}    & \textbf{57.83$\pm$1.01}                     \\
                          & ARI    & 49.55$\pm$0.37 & 48.44$\pm$0.14 & 51.26$\pm$0.21 & 19.61$\pm$0.22 & 26.18$\pm$0.36 & 31.12$\pm$0.18 & 24.50$\pm$0.40 & {\underline{ 55.36$\pm$0.37}}    & \textbf{60.55$\pm$1.78}                     \\
                          & F1     & 60.96$\pm$0.22 & 64.25$\pm$0.22 & 63.21$\pm$0.12 & 43.53$\pm$0.42 & 57.14$\pm$0.17 & 61.82$\pm$0.13 & 51.10$\pm$0.20 & {\underline{ 65.48$\pm$0.08}}    & \textbf{66.16$\pm$0.64}                     \\ \hline
\multirow{4}{*}{ACM}      & ACC    & 81.83$\pm$0.08 & 84.33$\pm$0.76 & 85.12$\pm$0.52 & 84.52$\pm$1.44 & 84.13$\pm$0.22 & 86.94$\pm$2.83 & 86.10$\pm$1.20 & {\underline{ 90.45$\pm$0.18} }   & \textbf{90.59$\pm$0.15}                     \\
                          & NMI    & 49.30$\pm$0.16 & 54.54$\pm$1.51 & 56.61$\pm$1.16 & 55.38$\pm$1.92 & 53.20$\pm$0.52 & 56.18$\pm$4.15 & 55.70$\pm$1.40 & {\underline{ 68.31$\pm$0.25}}    & \textbf{68.38$\pm$0.45}                     \\
                          & ARI    & 54.64$\pm$0.16 & 60.64$\pm$1.87 & 62.16$\pm$1.50 & 59.46$\pm$3.10 & 57.72$\pm$0.67 & 59.35$\pm$3.89 & 62.90$\pm$2.10 & {\underline{ 73.91$\pm$0.40}}    & \textbf{74.20$\pm$0.38}                     \\
                          & F1     & 82.01$\pm$0.08 & 84.51$\pm$0.74 & 85.11$\pm$0.48 & 84.65$\pm$1.33 & 84.17$\pm$0.23 & 87.07$\pm$2.79 & 86.10$\pm$1.20 & {\underline{ 90.42$\pm$0.19}}    & \textbf{90.58$\pm$0.17}                     \\ \hline
\multirow{4}{*}{CiteSeer} & ACC    & 57.08$\pm$0.13 & 55.89$\pm$0.20 & 60.49$\pm$1.42 & 61.35$\pm$0.80 & 60.97$\pm$0.36 & 64.54$\pm$1.39 & 56.90$\pm$0.70 & {\underline{ 65.96$\pm$0.31}}    & \textbf{68.79$\pm$0.23}                     \\
                          & NMI    & 27.64$\pm$0.08 & 28.34$\pm$0.30 & 27.17$\pm$2.40 & 34.63$\pm$0.65 & 32.69$\pm$0.27 & 36.41$\pm$0.86 & 34.50$\pm$0.80 & {\underline{ 38.71$\pm$0.32}}    & \textbf{41.54$\pm$0.30}                     \\
                          & ARI    & 29.31$\pm$0.14 & 28.12$\pm$0.36 & 25.70$\pm$2.65 & 33.55$\pm$1.18 & 33.13$\pm$0.53 & 37.78$\pm$1.24 & 33.40$\pm$1.50 & {\underline{ 40.17$\pm$0.43}}    & \textbf{43.79$\pm$0.31}                     \\
                          & F1     & 53.80$\pm$0.11 & 52.62$\pm$0.17 & 61.62$\pm$1.39 & 57.36$\pm$0.82 & 57.70$\pm$0.49 & 62.20$\pm$1.32 & 54.80$\pm$0.80 & \textbf{63.62$\pm$0.24} & {\underline{ 62.37$\pm$0.21}}                        \\ \hline
\multirow{4}{*}{DBLP}     & ACC    & 51.43$\pm$0.35 & 58.16$\pm$0.56 & 60.31$\pm$0.62 & 61.21$\pm$1.22 & 58.59$\pm$0.06 & 62.05$\pm$0.48 & 61.60$\pm$1.00 & {\underline{ 68.05$\pm$1.81}}    & \textbf{73.26$\pm$0.37}                     \\
                          & NMI    & 25.40$\pm$0.16 & 29.51$\pm$0.28 & 31.17$\pm$0.50 & 30.80$\pm$0.91 & 26.92$\pm$0.06 & 32.49$\pm$0.45 & 26.80$\pm$1.00 & {\underline{ 39.50$\pm$1.34} }   & \textbf{39.68$\pm$0.42}                     \\
                          & ARI    & 12.21$\pm$0.43 & 23.92$\pm$0.39 & 25.37$\pm$0.60 & 22.02$\pm$1.40 & 17.92$\pm$0.07 & 21.03$\pm$0.52 & 22.70$\pm$0.30 & {\underline{ 39.15$\pm$2.01}}    & \textbf{42.49$\pm$0.31}                     \\
                          & F1     & 52.53$\pm$0.36 & 59.38$\pm$0.51 & 61.33$\pm$0.56 & 61.41$\pm$2.23 & 58.69$\pm$0.07 & 61.75$\pm$0.67 & 61.80$\pm$0.90 & {\underline{ 67.71$\pm$1.51}}    & \textbf{72.80$\pm$0.56}                     \\ \hline\hline
\end{tabular}
}
\end{table*}

\begin{table*}[]
\setlength{\abovecaptionskip}{2pt}%    
\setlength{\abovecaptionskip}{8pt}%
\caption{The ablation study on six benchmark datasets. `AGCN-S[S]' indicates the AGCN-S module without the attention-based scale-wise mechanism, i.e., all the weights of the multi-scale features are set as $1$. `AGCN-S[A]' indicates the use of the corresponding attention-based scale-wise mechanism. `$\checkmark$' in each row denotes the usage of the corresponding component. The best results are highlighted with \textbf{bold}.}
\label{tab: AS}
\setlength{\tabcolsep}{1.48mm}{
\begin{tabular}{l|ccc|cccc}
\hline\hline
Datasets   & AGCN-S[A]      & AGCN-S[S]       & AGCN-H       & ACC                     & NMI                     & ARI                     & F1                      \\ \hline
\multirow{4}{*}{USPS}     &              &              &              & 78.08$\pm$0.19          & 79.51$\pm$0.27          & 71.84$\pm$0.24          & 76.98$\pm$0.18          \\ \cline{2-4}
                          &              &              & $\checkmark$ & 79.43$\pm$1.10          & 79.13$\pm$0.51          & 72.00$\pm$1.16          & 76.71$\pm$0.78          \\ \cline{2-4}
                          &              & $\checkmark$ & $\checkmark$ & 80.20$\pm$0.75          & 79.38$\pm$0.28          & 72.79$\pm$0.74          & 77.10$\pm$0.50          \\ \cline{2-4}
                          & $\checkmark$ & $\checkmark$ & $\checkmark$ & \textbf{80.98$\pm$0.28} & \textbf{79.64$\pm$0.32} & \textbf{73.61$\pm$0.43} & \textbf{77.61$\pm$0.38} \\ \hline
\multirow{4}{*}{HHAR}     &              &              &              & 84.26$\pm$0.17          & 79.90$\pm$0.09          & 72.84$\pm$0.09          & 82.58$\pm$0.08          \\ \cline{2-4}
                          &              &              & $\checkmark$ & 84.39$\pm$1.61          & 80.63$\pm$0.65          & 73.40$\pm$0.64          & 82.67$\pm$2.43          \\ \cline{2-4}
                          &              & $\checkmark$ & $\checkmark$ & 82.85$\pm$1.60          & 80.24$\pm$0.43          & 72.41$\pm$0.64          & 80.32$\pm$2.43          \\ \cline{2-4}
                          & $\checkmark$ & $\checkmark$ & $\checkmark$ & \textbf{88.11$\pm$0.43} & \textbf{82.44$\pm$0.62} & \textbf{77.07$\pm$0.66} & \textbf{88.00$\pm$0.53} \\ \hline
\multirow{4}{*}{Reuters}  &              &              &              & 77.15$\pm$0.21          & 50.82$\pm$0.21          & 55.36$\pm$0.37          & 65.48$\pm$0.08          \\ \cline{2-4}
                          &              &              & $\checkmark$ & 77.81$\pm$0.89          & 53.94$\pm$1.08          & 56.83$\pm$1.53          & 65.10$\pm$0.65          \\ \cline{2-4}
                          &              & $\checkmark$ & $\checkmark$ & 78.15$\pm$0.67          & 53.90$\pm$1.31          & 56.53$\pm$1.92          & 64.84$\pm$0.58          \\ \cline{2-4}
                          & $\checkmark$ & $\checkmark$ & $\checkmark$ & \textbf{79.30$\pm$1.07} & \textbf{57.83$\pm$1.01} & \textbf{60.55$\pm$1.78} & \textbf{66.16$\pm$0.64} \\ \hline
\multirow{4}{*}{ACM}      &              &              &              & 90.45$\pm$0.18          & 68.31$\pm$0.25          & 73.91$\pm$0.40          & 90.42$\pm$0.19          \\ \cline{2-4}
                          &              &              & $\checkmark$ & 90.47$\pm$0.24          & 68.42$\pm$0.61          & 73.95$\pm$0.60          & 90.48$\pm$0.26          \\ \cline{2-4}
                          &              & $\checkmark$ & $\checkmark$ & 90.57$\pm$0.11          & 68.43$\pm$0.42          & 74.16$\pm$0.30          & 90.56$\pm$0.11          \\ \cline{2-4}
                          & $\checkmark$ & $\checkmark$ & $\checkmark$ & \textbf{90.59$\pm$0.15} & \textbf{68.38$\pm$0.45} & \textbf{74.20$\pm$0.38} & \textbf{90.58$\pm$0.17} \\ \hline
\multirow{4}{*}{CiteSeer} &              &              &              & 65.96$\pm$0.31          & 38.71$\pm$0.32          & 40.17$\pm$0.43          & \textbf{63.62$\pm$0.24} \\ \cline{2-4}
                          &              &              & $\checkmark$ & 66.38$\pm$1.72          & 39.07$\pm$1.52          & 40.93$\pm$1.78          & 60.91$\pm$0.81          \\ \cline{2-4}
                          &              & $\checkmark$ & $\checkmark$ & 68.34$\pm$0.32          & 41.10$\pm$0.43          & 43.27$\pm$0.53          & 62.00$\pm$0.35          \\ \cline{2-4}
                          & $\checkmark$ & $\checkmark$ & $\checkmark$ & \textbf{68.79$\pm$0.23} & \textbf{41.54$\pm$0.30} & \textbf{43.79$\pm$0.31} & 62.37$\pm$0.21          \\ \hline
\multirow{4}{*}{DBLP}     &              &              &              & 68.05$\pm$1.81          & 39.50$\pm$1.34          & 39.15$\pm$2.01          & 67.71$\pm$1.51          \\ \cline{2-4}
                          &              &              & $\checkmark$ & 69.65$\pm$1.43          & 35.37$\pm$1.58          & 37.78$\pm$1.85          & 68.69$\pm$1.65          \\ \cline{2-4}
                          &              & $\checkmark$ & $\checkmark$ & 71.49$\pm$0.52          & 37.38$\pm$0.65          & 39.91$\pm$0.78          & 71.02$\pm$0.60          \\ \cline{2-4}
                          & $\checkmark$ & $\checkmark$ & $\checkmark$ & \textbf{73.26$\pm$0.37} & \textbf{39.68$\pm$0.42} & \textbf{42.49$\pm$0.31} & \textbf{72.80$\pm$0.56} \\ 
\hline\hline
\end{tabular}
}
\end{table*}

\begin{itemize}
	\item	\textbf{USPS}. The United States Postal Service database includes ten classes (i.e., `0'–`9') of 11000 handwritten digits. We use a popular subset containing 9298 handwritten digit images for the experiments, and all of these images are normalized to $16 \times 16$.
	\item	\textbf{HHAR}. The Heterogeneity Human Activity Recognition database contains 10299 sensor records from smartphones and smartwatches. All samples are partitioned into 6 categories of human activities, including: `Biking', `Sitting', `Standing', `Walking', `Stair Up' and `Stair Down'.
	\item	\textbf{Reuters}. The Reuters dataset is a collection of English news, labeled by category. We use four root categories: corporate/industrial, government/social, markets, and economics as labels and sample a random subset of 10000 examples for clustering.
	\item 	\textbf{ACM}. The ACM dataset is a paper network from ACM digital library, in which two papers are connected with an edge if they are written by the same author. The features are selected from KDD, SIGMOD, SIGCOMM, MobiCOMM keywords with three classes (i.e., database, wireless communication, data mining) by their research area.
	\item	\textbf{CiteSeer}. The CiteSeer is a citation network containing sparse bag-of-words feature vectors for each document and a list of citation links between documents. The labels contain six areas: agents, artificial intelligence, database, information retrieval, machine language, and human-computer interaction.
	\item	\textbf{DBLP}. The DBLP dataset is an author network from the dblp computer science bibliography, in which two authors are connected with an edge if they have the coauthor relationship. The author features are the elements of a bag-of-words represented of keywords, in which authors are divided into four areas: database, data mining, machine learning, and information retrieval and labeled according to the conferences they submitted.
\end{itemize}

\subsection{Compared Methods} 

We compare our method with three types of methods, including AE-based clustering methods \cite{hinton2006reducing,xie2016unsupervised,guo2017improved}, attention-based clustering method \cite{wang2019attributed}, and GCN-based clustering methods \cite{kipf2016variational,8822591,bo2020structural}:

\begin{itemize}
	\item	\textbf{AE} performs K-means \cite{macqueen1967some} on the deep representations learned by the auto-encoder module \cite{hinton2006reducing}.
	\item	\textbf{DEC} \cite{xie2016unsupervised} clusters a set of data points in a jointly optimized feature space.
	\item	\textbf{IDEC} \cite{guo2017improved} is a variant of DEC by adding a reconstruction loss.
	\item	\textbf{GAE} and \textbf{VGAE} \cite{kipf2016variational} use GCN to learn data representations in an unsupervised graph embedding manner based on AE and variational AE frameworks, respectively. 
	\item	\textbf{DAEGC} \cite{wang2019attributed} uses the attentional neighbor-wise strategy to learn the node representations and employs a clustering loss to supervise the process of graph clustering.
	\item   \textbf{ARGA} \cite{8822591} develops an adversarial regularizer to guide the learning of latent representations.
	\item	\textbf{SDCN} \cite{bo2020structural} integrates the structural information into deep clustering via the combination of DEC and GCN.
\end{itemize}

\subsection{Implementation Details}
\textbf{Training Procedure:}
For fair comparisons, we follow the same network parameter settings as \cite{xie2016unsupervised,guo2017improved,bo2020structural}, i.e., the dimension of the auto-encoder is set to $500-500-2000-10$. Furthermore, the dimension of the GCN layers is also set to $500-500-2000-10$. The training of our AGCN method includes two phases. In the first phase, we pre-train the AE module with $30$ epochs and the learning rate is set to $0.001$. In the second phase, the whole network is trained for $200$ iterations (i.e., $i_\emph{MaxIter}=200$). The learning rates of USPS, HHAR, ACM, and DBLP datasets are set to $0.001$, and the learning rates of Reuters and CiteSeer datasets are set to $0.0001$. $\lambda_1$ and $\lambda_2$ are set to $\left\{1000, 1000\right\}$ for USPS, $\left\{1, 0.1\right\}$ for HHAR, $\left\{10, 10\right\}$ for Reuters, and $\left\{0.1, 0.01\right\}$ for graph datasets. The batch size of the network is set to $256$. For the ARGA method, we conduct the parameter settings given by the original paper \cite{8822591}. For other comparisons, we directly cite the results in \cite{bo2020structural}. Following all the compared methods, we repeat the experiment 10 times to evaluate our method and report the mean values and the corresponding standard deviations (i.e., mean$\pm$std). The training procedure is implemented with PyTorch and a GPU (GeForce RTX 2080 Ti). The code will be publicly available upon acceptance.

\textbf{Evaluation Metrics:}
To evaluate the clustering performance of all the methods, we use four metrics, including Accuracy (ACC), Normalized Mutual Information (NMI), Average Rand Index (ARI), and macro F1-score (F1). For each metric, a larger value implies a better clustering result. The detailed definitions of those metrics can be found in \cite{bo2020structural}.

\subsection{Clustering Results}

The experimental results of our method and eight compared methods on six benchmark datasets are shown in Table \ref{tab: final_result}, in which the bold values and the underlined values indicate the best and second-best clustering performances, respectively. As shown in Table \ref{tab: final_result}, we have the following observations:
\begin{itemize}
	\item Our method obtains the best clustering performance among all the comparisons in most circumstances. For example, in the \textbf{non-graph} dataset HHAR, our approach improves 3.85\% over the second-best comparison on ACC, 2.54\% on NMI, 4.23\% on ARI, and 5.42\% on F1 averagely. In addition, in the \textbf{graph} dataset DBLP, our approach improves 5.21\% over the second-best comparison on ACC, 0.18\% on NMI, 3.34\% on ARI, and 5.09\% on F1 averagely. The reason for the significant improvement is three-fold. First, our method adaptively fuses the GCN feature and the AE feature for exploiting the numerous and discriminative information as far as possible. Second, our approach dynamically combines the multi-scale features to make full use of the information of each layer. Last but not least, our designed training strategy can develop more robust guidance for clustering by providing abundant and discriminative information to construct the soft assignment.
	\item DAEGC performs better than GAE, validating the importance of considering the attention-based mechanism. By extending the attention-based mechanism to the heterogeneity-wise and scale-wise feature fusions, our AGCN-H and AGCN-S modules are capable of making a further and significant performance improvement.
	\item SDCN performs better than the AE-based clustering methods (AE, DEC, IDEC) and the GCN-based methods (GAE, VGAE, ARGA), validating the importance of combing AE and GCN models together. However, SDCN equates the importance between the graph structure feature and the node attribute feature and neglects the multi-scale features, resulting in the sub-optimal clustering performance. By solving the aforementioned drawbacks, our approach is capable of gaining the best results in all six datasets.
	\item In the ACM dataset, the performance improvement of our method is not significant. The reason is possible that in the graph of ACM, many nodes are already well-connected, making a prominent clustering performance even with one GCN layer. However, many real-world applications do not owe a good graph. For example, the graph quality of Reuters is not high, resulting a relatively low clustering performance for the GCN-based methods. In this case, exploiting the dynamic feature fusion strategy and considering the multi-scale features information are essential to improve the clustering performance, e.g., the performance improvement of our method in Reuters is significant.
\end{itemize}

\begin{figure}
    \centering
	\includegraphics[width=0.56\columnwidth] {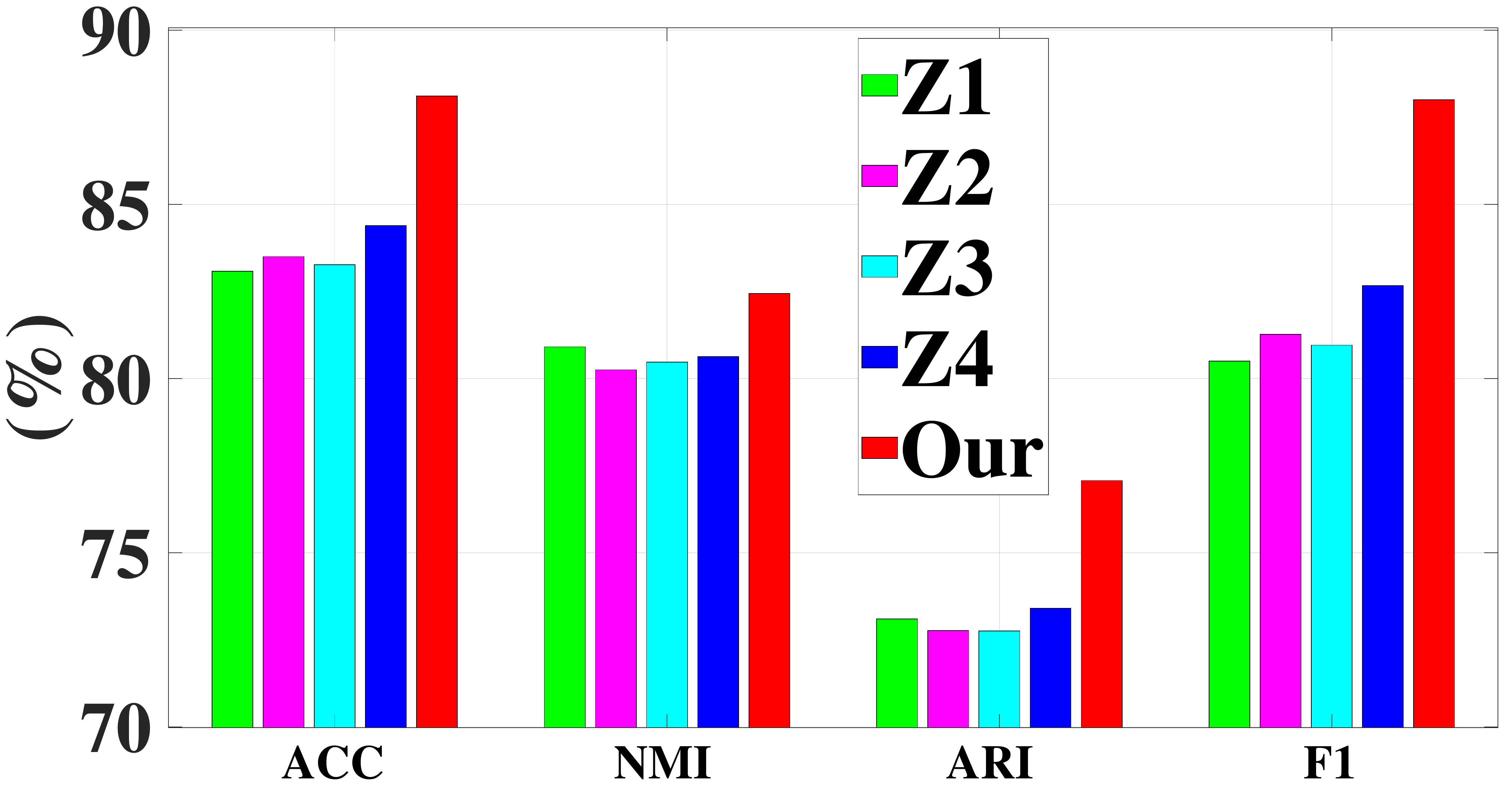} 
    \setlength{\abovecaptionskip}{2pt}%    
    \setlength{\abovecaptionskip}{8pt}%
    \caption{Analysis of different scale layers on the HHAR dataset. \textbf{`Z1', `Z2', `Z3',} and \textbf{`Z4'} denote the cases of using each scale feature as the input, respectively; `\textbf{Our}' denotes our case of using the fused feature as the input of the prediction layer. Here, the ordinate indicates the mean results.}
    \vspace{-0.3cm}
    \label{fig: AS-scale}
\end{figure}

\begin{figure}
    \centering
	\subfigure[USPS]{
	\includegraphics[width=0.30\columnwidth] {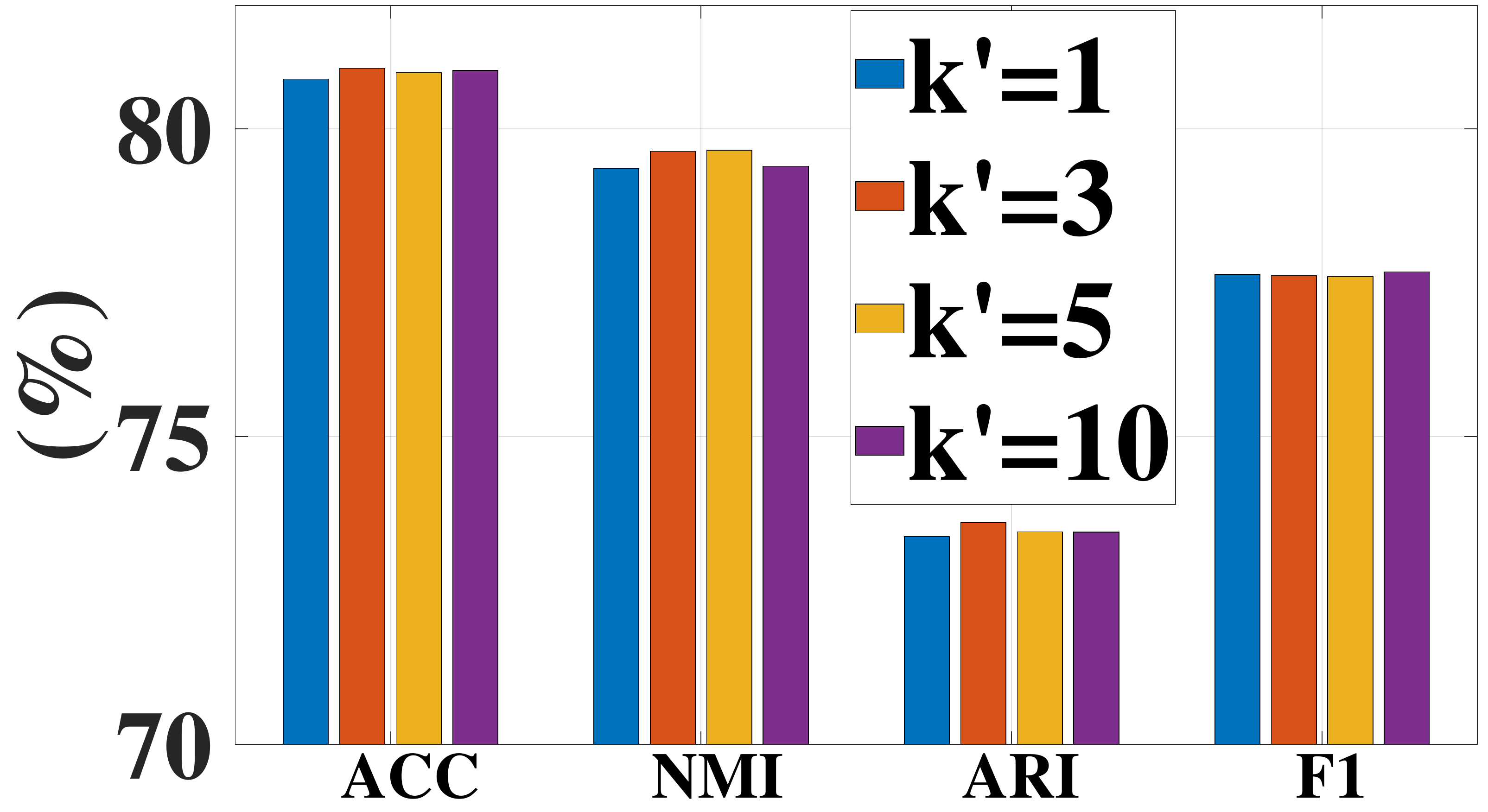}
	}
	\subfigure[HHAR]{
	\includegraphics[width=0.30\columnwidth] {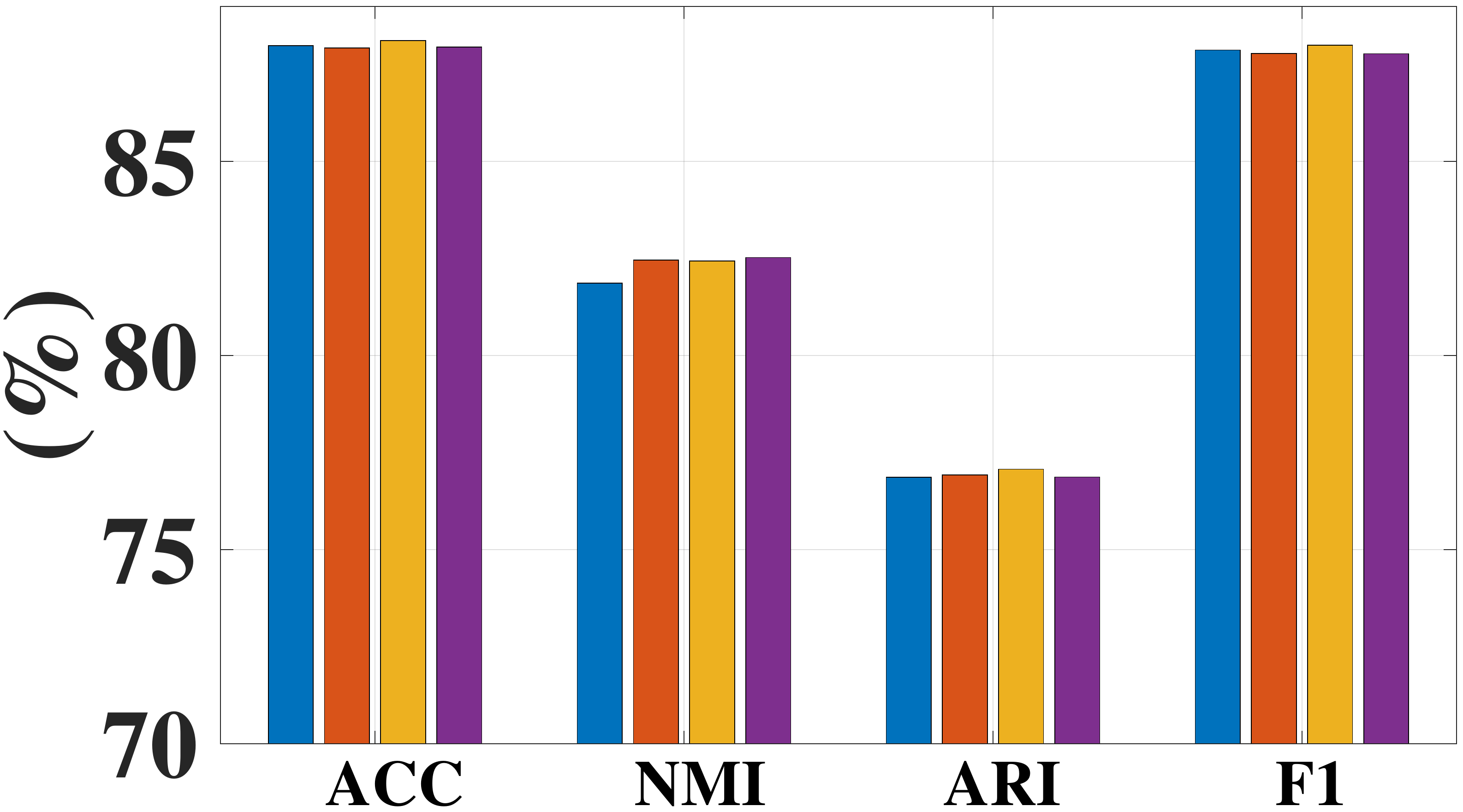} 
	}
	\subfigure[Reuters]{
	\includegraphics[width=0.30\columnwidth] {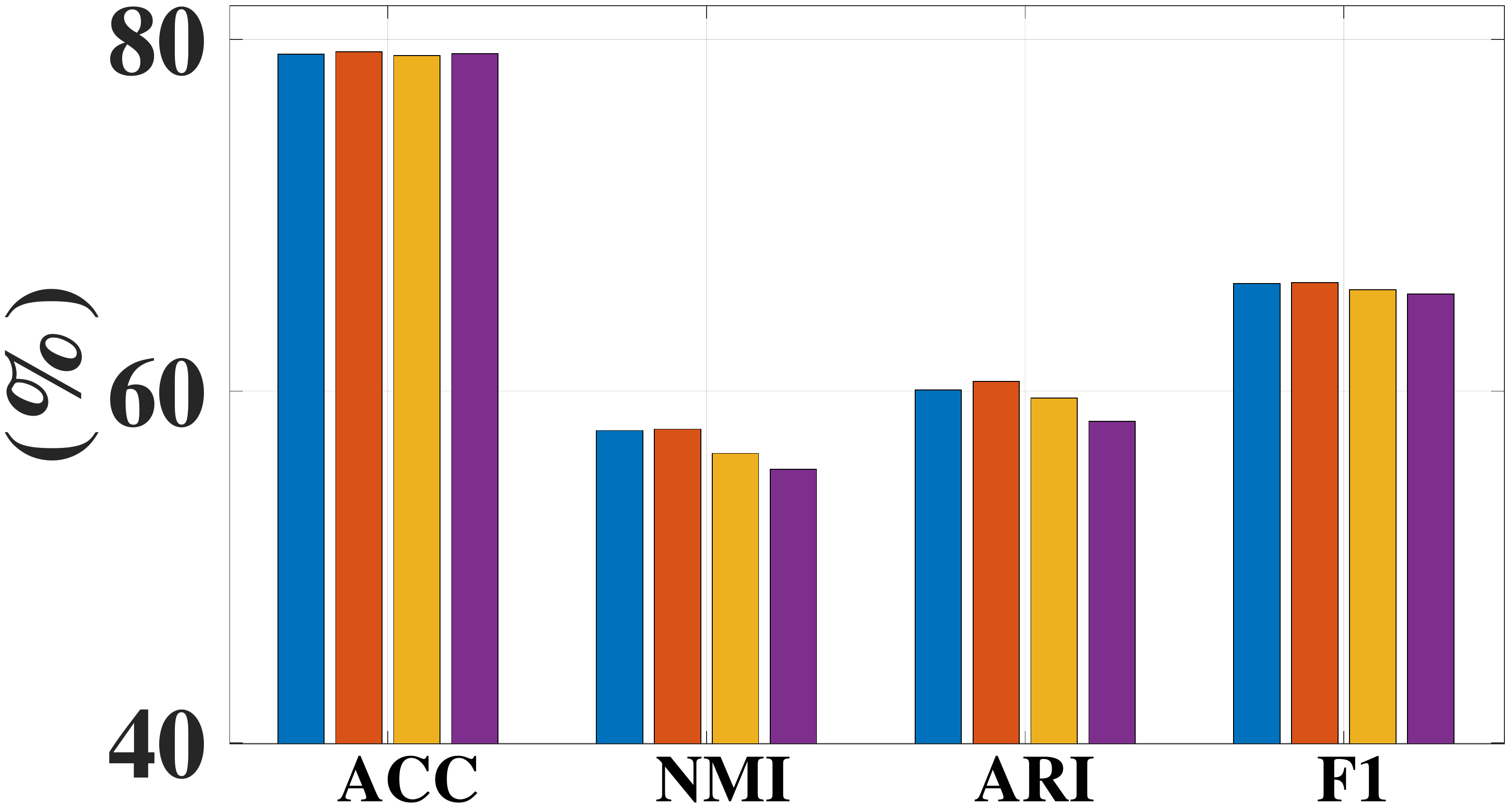} 
	}
	\setlength{\abovecaptionskip}{2pt}%    
    \setlength{\abovecaptionskip}{8pt}%
    \caption{Clustering results with different $k'$.}
    \vspace{-0.3cm}
    \label{fig: AS-k}
\end{figure}

\subsection{Ablation Study}

We conduct ablation studies to evaluate the efficiency and effectiveness of the AGCN-H module and the AGCN-S module. Besides, we also analyze the influence of different scale features on the clustering performance. The results are reported in Table \ref{tab: AS}.

\textbf{Analysis of AGCN-H module.}
We start by examining the AGCN-H module, in which the experimental comparisons are shown in the first row (without the AGCN-H module) and the second row (with the AGCN-H module) of each dataset in Table \ref{tab: AS}. 
We can observe that the AGCN-H module produces performance improvement to a certain extent, which validates the effectiveness of the attention-based heterogeneity-wise strategy, i.e., learning a flexible representation with the dynamic weighted mechanism is conducive to obtain better clustering results.

\textbf{Analysis of AGCN-S module.}
We evaluate the AGCN-S module from two aspects, including ($i$) the multi-scale feature fusion (marked as AGCN-S[S]) and ($ii$) the attention-based scale-wise strategy (marked as AGCN-S[A]). 

\begin{itemize}
    \item For the first aspect, by comparing the experimental results shown in the second and third rows of each dataset in Table \ref{tab: AS}, we can find that the multi-scale feature fusion can help obtain better clustering performance in most cases. The only exception is HHAR where some features of the middle layers suffer from the over-smoothing issue, resulting in the negative propagation.
    \item For the second aspect, by comparing each dataset results of the third and fourth row in Table \ref{tab: AS}, we can find that considering the attention-based scale-wise strategy is capable of obtaining the best clustering performance. Especially, in the HHAR dataset, considering the attention-based scale-wise strategy can sufficiently cope with the above-mentioned performance dropping. This phenomenon is credited to the fact that the attention-based scale-wise strategy can assign some negative features with a small weight value, avoiding the negative propagation. This once validates the effectiveness of the attention-based mechanism. 
\end{itemize}

\textbf{Analysis of different scale features.} 
To evaluate the contributions of different scale features to the clustering performance, we conduct clustering using different layers of the proposed model on the HHAR dataset. From Figure \ref{fig: AS-scale}, we can observe that dynamically fusing the features from different layers can significantly improve the clustering performance compared with the ones only using the feature from one layer. 

\textbf{Analysis of different $k'$.} 
As the number of neighbors $k'$ significantly influences the quality of the adjacency matrix, we conduct the parameter analysis of $k'$ on non-graph datasets, i.e., USPS, HHAR, and Reuters. From Figure \ref{fig: AS-k}, we can observe that our model is not sensitive to $k'$. 

\begin{figure}
    \centering
    \includegraphics[width=0.8\columnwidth] {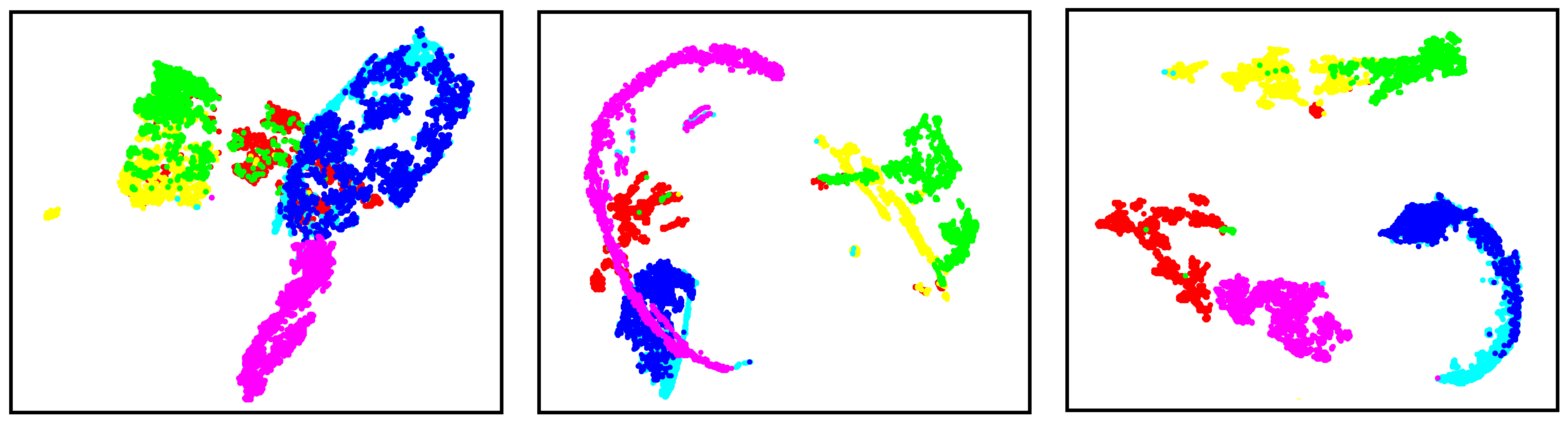} 
    \setlength{\abovecaptionskip}{2pt}%    
    \setlength{\abovecaptionskip}{8pt}%
    \caption{Visualization comparison of embeddings from raw data, the second-best comparison \textcolor{black}{(SDCN)}, and our method (from left to right) on the HHAR dataset. The different colors represent different groups.}
    \vspace{-0.3cm}
    \label{fig: tsne}
\end{figure}

\subsection{Visualization} 
To intuitively verify the effectiveness of our method, we plot 2D t-distributed stochastic neighbor embedding (t-SNE) \cite{maaten2008visualizing} visualizations of the learned representations of our method as well as the best-compared ones on the HHAR dataset in Figure \ref{fig: tsne}. We can find that the feature representation obtained by our method shows the best separability for different clusters, where samples from the same class naturally gather together and the gap between different groups is the most obvious one. This phenomenon substantiates that our method produces the most discriminative representation compared with state-of-the-art methods.

\section{Conclusion}

In this paper, we proposed a novel deep clustering method termed Attention-driven Graph Clustering Network (AGCN) by simultaneously considering the dynamic fusion strategy and the multi-scale features fusion. By leveraging two novel attention-based fusion modules, AGCN is capable of adaptively learning the weights heterogeneity-wisely and scale-wisely for achieving those feature fusions. 
Moreover, extensive experiments on commonly used benchmark datasets validated the superiority of the proposed network over state-of-the-art methods, especially for the low-quality graph.

\bibliographystyle{ACM-Reference-Format}
\balance
\bibliography{sample-base}

\end{document}